\documentclass[letterpaper, 10 pt, conference]{ieeeconf}  

\IEEEoverridecommandlockouts                              

\usepackage{multirow}
\usepackage{graphicx}
\usepackage[utf8]{inputenc}
\usepackage[T1]{fontenc}
\usepackage{xcolor}
\usepackage{comment}
\usepackage{amssymb}
\usepackage{amsmath}
\usepackage{booktabs, multirow}
\usepackage{subcaption}
\usepackage{pifont}
\usepackage{array, makecell}
\usepackage{url}
\usepackage{mathrsfs}
\usepackage[colorlinks=true,linkcolor=cyan,urlcolor=red]{hyperref}
\usepackage{cite}
\hypersetup{
    colorlinks=true,
    linkcolor=black,
    filecolor=magenta,
    citecolor=black,
    urlcolor=black,
    }

\title{\LARGE \bf
Tri-MTL: A Triple Multitask Learning Approach for Respiratory Disease Diagnosis
}

\author{June-Woo Kim$^{1,2}$ Sanghoon Lee$^{3}$ Miika Toikkanen$^{2}$ Daehwan Hwang$^{4}$ Kyunghoon Kim$^{4,5}$$^{\dagger}$
\\ kaen2891@gmail.com \quad journey237@snu.ac.kr
\thanks{\line(1,0){150}}
\thanks{$^{\dagger}$Corresponding author}  
\thanks{*This work was supported by the New Faculty Startup Fund from Seoul National University (800-20230600), and was supported by Brian Impact Foundation, a non-profit organization dedicated to the advancement of science and technology for all.}
\thanks{$^{1}$Department of Psychiatry, Wonkwang University Hospital, Republic of Korea, $^{2}$RSC LAB, MODULABS, Republic of Korea $^{3}$The State University of New York, Korea, Republic of Korea $^{4}$Department of Pediatrics, Seoul National University Bundang Hospital, Republic of Korea. $^{5}$Department of Pediatrics, Seoul National University College of Medicine, Republic of Korea.}
}

\begin{document}

\newcommand{\jw}[1]{{\color{orange} #1}}

\maketitle
\thispagestyle{empty}
\pagestyle{empty}

\begin{abstract}
Auscultation remains a cornerstone of clinical practice, essential for both initial evaluation and continuous monitoring. Clinicians listen to the lung sounds and make a diagnosis by combining the patient's medical history and test results. 
Given this strong association, multitask learning (MTL) can offer a compelling framework to simultaneously model these relationships, integrating respiratory sound patterns with disease manifestations. While MTL has shown considerable promise in medical applications, a significant research gap remains in understanding the complex interplay between respiratory sounds, disease manifestations, and patient metadata attributes. This study investigates how integrating MTL with cutting-edge deep learning architectures can enhance both respiratory sound classification and disease diagnosis. Specifically, we extend recent findings regarding the beneficial impact of metadata on respiratory sound classification by evaluating its effectiveness within an MTL framework. Our comprehensive experiments reveal significant improvements in both lung sound classification and diagnostic performance when the stethoscope information is incorporated into the MTL architecture.

\indent \textit{Clinical relevance}— Our integrated MTL approach has immediate clinical applications in supporting medical professionals' diagnostic decisions, including lung sound classification to aid in detecting respiratory disorders, potentially reducing misdiagnosis rates and improving patient outcomes in respiratory care settings (85.83\% and 78.86\% specificity, along with 94.09\% and 41.56\% sensitivity, for disease diagnosis and lung sound classification, respectively.).
\end{abstract}

\section{INTRODUCTION}
Respiratory diseases pose a significant global health challenge, impacting millions and placing substantial strain on healthcare systems. Among the primary diagnostic methods, auscultation remains a cornerstone of clinical practice, essential for both initial evaluation and continuous monitoring of respiratory pathologies~\cite{coucke2019laennec}. 
As a non-invasive and real-time diagnostic tool, the stethoscope remains essential in modern medicine~\cite{sarkar2015auscultation}. It plays a particularly vital role in diagnosing respiratory diseases, as abnormal respiratory sounds often reveal critical insights into pathological conditions of the lungs and bronchi~\cite{kim2022coming}. 


Clinicians generally employ auscultation to detect abnormal respiratory sounds, which are crucial indicators of underlying pathological conditions. These acoustic signatures, including wheezes, crackles, and rhonchi, demonstrate strong correlations with specific respiratory diseases, making them valuable diagnostic markers~\cite{miller2007approach, sarkar2015auscultation}. 
However, they do not conclude the diagnosis based on lung sounds alone, but rather combine the lung sounds, patient history, and test results to make the decision.
This diagnostic approach underscores the importance of exploring multitask learning (MTL) frameworks in the deep-learning domain, which can simultaneously handle multiple related diagnostic tasks~\cite{chaichulee2017multi, yang2017novel, li2019canet, liao2019multi, amyar2020multi, gao2020feature, song2020end, suma2024multi, wang2024multi, kim2023cross, kim2023intra}. MTL can capture the complex relationships between various aspects of medical data, enhancing the robustness of the model and improving the performance of each task while optimizing diagnostic workflows~\cite{vandenhende2020revisiting, kv2024multi, cui2024automated}. Despite these advances, a crucial gap exists in understanding how respiratory sounds, disease manifestations, and metadata interact within an MTL framework.


While recent studies have demonstrated that metadata can enhance the accuracy of respiratory sound classification~\cite{moummad2023pretraining, kim2024stethoscope, kim24f_interspeech}, the potential of incorporating this information within an MTL framework remains largely unexplored. This integration is particularly relevant as patient characteristics and other metadata attributes provide valuable context for interpreting respiratory sounds~\cite{moummad2023pretraining, kim2024stethoscope, kim24f_interspeech, kim2025adaptive} and improving diagnostic accuracy.

In this paper, we address these gaps by investigating the potential of an integrated MTL approach that combines respiratory sound analysis and disease diagnosis tasks with patients' metadata. To this end, we expand on recent findings highlighting the beneficial impact of metadata attributes on respiratory sound classification by integrating these insights into the MTL framework using a state-of-the-art pretrained Audio Spectrogram Transformer (AST) model~\cite{gong2021ast, bae23b_interspeech}. Experimental results on the ICBHI~\cite{rocha2018alpha} dataset reveal that incorporating electronic stethoscope metadata into the MTL framework notably improves both lung sound classification and diagnostic performance.
The primary contributions of this paper are threefold:
\begin{itemize}
    \item We propose a simple yet effective MTL framework that integrates respiratory sound analysis and disease diagnosis tasks with recent state-of-the-art AST~\cite{gong2021ast} model, resulting in improved diagnostic accuracy.
    \item We introduce \textbf{Tri-MTL} (lung sound + disease + metadata attributes), an MTL framework incorporating lung sounds, disease labels, and metadata attributes, building on recent findings that demonstrate the benefits of metadata in respiratory sound classification.
    \item We provide comprehensive experimental evidence supporting the effectiveness of incorporating stethoscope metadata attributes within the MTL architecture.
\end{itemize}

\section{METHODS: MULTITASK LEARNING}
\begin{figure*}[ht!]
    \centering
    \includegraphics[width=1.0\linewidth]{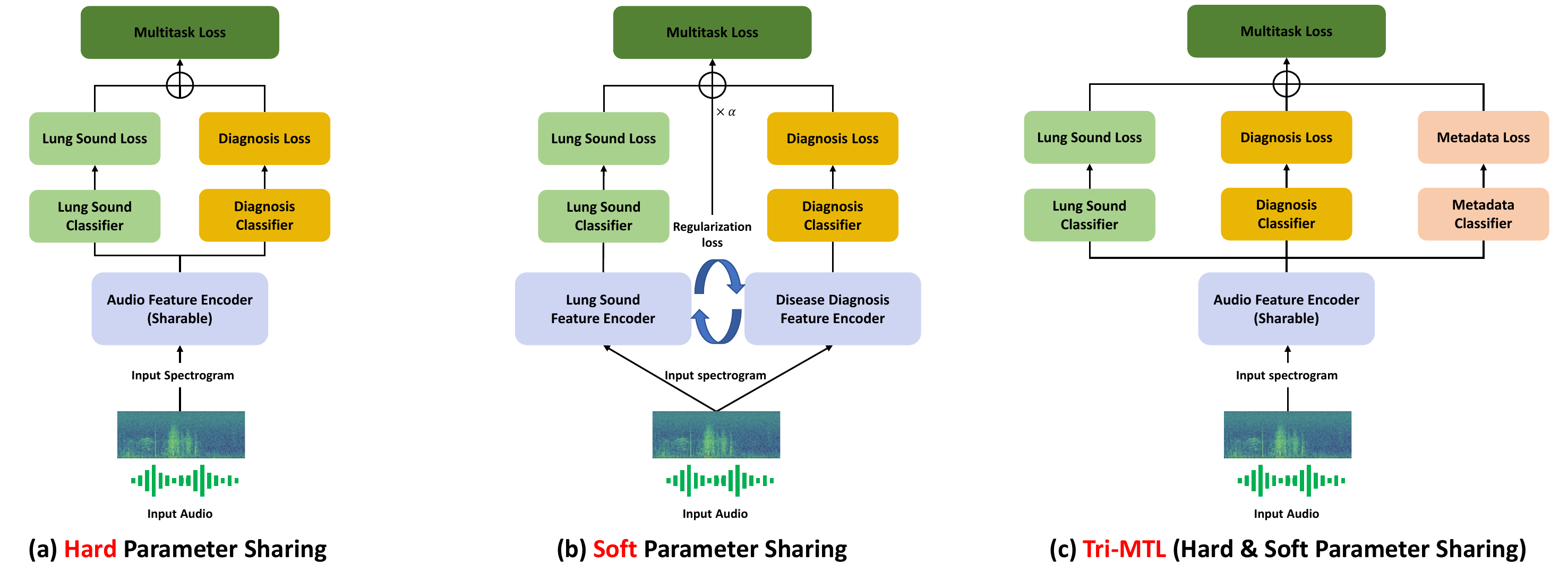}
    \caption{Illustration of the proposed MTL approaches for joint lung sound classification and disease diagnosis. 
    {$(a)$ Hard parameter sharing utilizes a single shared audio feature encoder to extract representations for both tasks, followed by separate classifiers for lung sound classification and disease diagnosis. $(b)$ Soft parameter sharing introduces two task-specific encoders, one for lung sound classification and the other for disease diagnosis, with a regularization loss encouraging shared representation learning between them. $(c)$ Tri-MTL extends both lung sound classification and disease diagnosis tasks by incorporating an additional metadata classification task, enhancing the model’s ability to leverage metadata attributes. For the Tri-MTL, both hard and soft parameter sharing strategies are considered.}
    }
    \label{fig:fig1}
\end{figure*}

MTL (multitask learning) offers numerous advantages, including reduced redundant computations, lower memory requirements, and faster inference times, all of which contribute to improved training efficiency~\cite{ruder2017overview}. Moreover, task-sharing within a unified network enables the exchange of complementary information, acting as a regularizer that mitigates overfitting~\cite{baxter1997bayesian} and therefore improves model performance. Accordingly, we present three MTL approaches as described in Figure~\ref{fig:fig1}: \textbf{hard parameter sharing} and \textbf{soft parameter sharing} for both lung sound classification and disease diagnosis, as well as \textbf{Tri-MTL}, which additionally integrates metadata attributes.

\subsection{Hard parameter sharing}
Hard parameter sharing is one of the most widely used approaches for MTL in deep learning~\cite{caruana1993multitask}. Given an input lung sound $\mathbf{x}$, all tasks share a common audio feature extractor $f_{\text{shared}}$, which extracts a shared representation $\mathbf{h}$:
\begin{equation}
    \mathbf{h} = f_{\text{shared}}(\mathbf{x};\theta_{\textbf{shared}})
\label{equ1}
\end{equation}
where $f_{\text{shared}}$ is the shared audio encoder and $\theta_{\textbf{shared}}$ are the shared parameters of $f_{\text{shared}}$. Each task $t$ then has a task-specific head (classifier) $g_t$, which maps the shared representation to a task-specific output:
\begin{equation}
    \hat{\mathbf{y}}_t = g_t(\mathbf{h};\theta_{g_t})
\label{equ2}
\end{equation}
where $\theta_{g_t}$ represents the parameters of the task-specific head. Each task has an individual loss function $\mathcal{L}_t$, and the final loss of the hard parameter sharing is simply computed as the sum of all task losses:
\begin{equation}
    {\mathcal{L}}_\text{Hard} = \! = \sum_{t=1}^T\! \, \mathcal{L}_{t}\! \,(\hat{\mathbf{y}}_t, \mathbf{y}_t)
\label{equ3}
\end{equation}
where $\mathcal{L}_{t}(\hat{\mathbf{y}}_t, \mathbf{y}_t)$ is the loss for task $t$ and $\mathbf{y}_t$ is the ground truth for task $t$. Since the audio feature encoder $f_\text{shared}$ is shared, the gradients from different tasks influence the same shared parameters.

For both lung sound classification and disease diagnosis tasks, Equation (\ref{equ2}) can be expressed as $\hat{\mathbf{y}}_l=g_l(\mathbf{h};\theta_l)$ for lung sound classification and $\hat{\mathbf{y}}_d=g_d(\mathbf{h};\theta_d)$ for disease diagnosis, where $l$ and $d$ denote the respective tasks. Therefore, Equation (\ref{equ3}) can be ${\mathcal{L}}_\text{Hard} = \mathcal{L}_\text{Lung} + \mathcal{L}_\text{Disease}$, and each lung sound classification loss and disease diagnosis loss can be formularized as:
\begin{equation}
    \mathcal{L}_{\text{Lung}} \! = -\sum_{i=1}^n\! \, y_{i}\! \, \log \, \!(\hat{y_{i}}), \quad \mathcal{L}_{\text{Disease}} \! = -\sum_{i=1}^n\! \, d_{i}\! \, \log \, \!(\hat{d_{i}}).
\label{equ4}
\end{equation}
where $\mathcal{L}_{\text{Lung}}$ and $\mathcal{L}_{\text{Disease}}$ denote the Cross-Entropy loss for lung sound classification and disease diagnosis, respectively, with ground truth labels $y$ and $d$ (omitting division by $N$). The predicted probabilities $\hat{y}$ and $\hat{d}$ are obtained by passing feature representations $\mathbf{h}$ through the lung sound classifier $g_l$ and the disease diagnosis classifier $g_d$, respectively.

\subsection{Soft parameter sharing}
In contrast to hard parameter sharing, soft parameter sharing allows each task-specific model to have independent parameters while adding a regularization loss to encourage similarity between models. Unlike Equation (\ref{equ2}), each task $t$ has an independent model or feature encoder $f_t$:
\begin{equation}
    \hat{\mathbf{y}}_t = f_t(\mathbf{x};\theta_{t})
\label{equ5}
\end{equation}
where $f_t$ denotes the task-specific feature encoder, and $\theta_t$ represents its corresponding parameters. The total loss for soft parameter sharing remains the same as in Equation (\ref{equ3}). As illustrated in Figure~\ref{fig:fig1}(b), the model employs two distinct feature encoders; one for lung sound classification and another for disease diagnosis.

To enforce parameter similarity between feature encoders, an L2 regularization loss~\cite{duong2015low} can be formularized as:
\begin{align}
    \mathcal{L}_{\text{pairwise}}(t, s) &= \sum_{l\in S} ||\theta_t^l - \theta_s^l||^2, \\
    \mathcal{L}_{\text{Reg}} &= \lambda \sum_{t=1}^{T} \sum_{s=t+1}^{T} \mathcal{L}_{\text{pairwise}}(t, s).
\label{equ6}
\end{align}
where $\lambda$ is the regularization hyperparameter, $T$ denotes the number of tasks, $\theta_t^l$ represents the weights of layer $l$ in task $t$, and $S$ is the set of layers where regularization is applied. The term $||\theta_t^l-\theta_s^l||^2$ measures the squared difference in weights between task $t$ and task $s$, where task $s$ corresponds to the $(t+1)$th task. This formulation ensures that all task pairs are penalized based on their weight differences. The total loss function of soft parameter sharing is given by $\mathcal{L}_\text{Soft}=\mathcal{L}_{\text{Task}}+\mathcal{L}_{\text{Reg}}$. The final loss terms for both lung sound classification with disease diagnosis tasks can be expanded as:
\begin{equation}
    \mathcal{L}_{\text{Soft}} \! = \mathcal{L_\text{Lung}} + \mathcal{L_\text{Disease}} + 
    \mathcal{L_\text{Reg}}
\label{equ7}
\end{equation}

\subsection{Tri-MTL: Integrating Metadata into Multitask Learning}
To further enhance the performance of respiratory disease diagnosis and lung sound classification, we propose \textbf{Tri-MTL}, an extended MTL framework that integrates metadata attributes alongside lung sound classification and disease diagnosis, as illustrated in Figure~\ref{fig:fig1}(c). Tri-MTL is strongly inspired by recent findings~\cite{kim2024stethoscope, kim24f_interspeech}, which demonstrated that incorporating or performing domain adaptation with patient metadata, such as age, gender, and stethoscope type, can significantly improve respiratory sound classification accuracy. Unlike conventional MTL frameworks that only leverage hard or soft parameter sharing between lung sound classification and disease diagnosis, Tri-MTL introduces a third task that explicitly models metadata information. By jointly optimizing lung sound classification, disease diagnosis, and metadata, Tri-MTL captures richer contextual information, enhancing the generalization of the model.

For hard parameter sharing on Tri-MTL, metadata classifier can be denoted as $\hat{\mathbf{y}}_m = g_m(\mathbf{h};\theta_m)$, and the total loss can be represented as:
\begin{equation}
    \mathcal{L_\text{Hard}}=\mathcal{L_\text{Lung}} + \mathcal{L_\text{Disease}} + \mathcal{L_\text{Meta}}
\label{equ8}
\end{equation}
where $m_i$ and $\hat{m_i}$ are ground truth metadata label and its corresponding predicted probability, $\mathcal{L}_{\text{Meta}} \! = -\sum_{i=1}^n\! \, m_{i}\! \, \log \, \!(\hat{m_{i}})$. For soft parameter sharing on Tri-MTL, the final loss can be formularized as:
\begin{equation}
    \mathcal{L}_{\text{Soft}} \! = \mathcal{L_\text{Lung}} + \mathcal{L_\text{Disease}} +
    \mathcal{L_\text{Meta}} + \mathcal{L_\text{Reg}}
\label{equ9}
\end{equation}

\section{EXPERIMENTS}
\subsection{Experimental Settings}
\subsubsection{Dataset}
We used the ICBHI respiratory dataset~\cite{rocha2018alpha}, which comprises approximately 5.5 hours of respiratory sound recordings. The dataset was officially pre-split into training (60\%) and testing (40\%) sets on breathing cycle level, ensuring no patient overlap between splits. Table~\ref{tab:tab1} provides a detailed breakdown of ICBHI dataset. We followed binarizing the age groups into adult (over 18 years old) and pediatric (18 years old or younger)~\cite{kim24f_interspeech, kim2025adaptive}. Other than the age group (i.e., sex, recording location, and recording stethoscope), we adhered to the official ICBHI metadata format.

\begin{table}[!t]
    \centering
    \caption{Overview of ICBHI dataset on the distribution of audio samples across lung sound and disease classes, along with various metadata types. L/R indicates whether the recording was taken from the left or right side.}
    \label{tab:tab1}
    \addtolength{\tabcolsep}{1pt}
    \resizebox{\linewidth}{!}{
    \begin{tabular}{clccc}
    \toprule
    & Label & Train & Test & Sum \\
    \hline \midrule
    \multirow{4}{*}{\multirow{4}{*}\textbf{Lung Sound}} & Normal & 2,063 & 1,579 & 3,642 \\
    & Crackle & 1,215 & 649 & 1,864 \\
    & Wheeze & 501 & 385 & 886 \\
    & Both & 363 & 143 & 506 \\
    \midrule

    \multirow{2}{*}{\multirow{2}{*}\textbf{Disease}} & Healthy & 147 & 175 & 322 \\
    & Unhealthy & 3,995 & 2,581 & 6,576 \\
    \midrule

    & Type & \multicolumn{3}{c}{Metadata Label} \\
    \midrule

    \multirow{5}{*}{\multirow{4}{*}\textbf{Metadata}} & Age & \multicolumn{3}{c}{Adult, Pediatric} \\
    & Sex & \multicolumn{3}{c}{Male, Female} \\
    & Location & \multicolumn{3}{c}{Trachea, L/R Anterior, L/R Posterior, L/R Lateral} \\
    & Stethoscope & \multicolumn{3}{c}{Meditron, LittC2SE, Litt3200, AKGC417L} \\
    \bottomrule
    \end{tabular}}
\end{table}
\begin{table*}[!t]
    \centering
    \caption{Performance comparison of different multitask learning (MTL) approaches. The table presents results for lung sound classification, disease diagnosis, and metadata prediction accuracy. Values are reported as mean $\pm$ standard deviation from five seeds. \textbf{Best} and {\underline{second best}} results.}
    \label{tab:tab2}
    \begin{tabular}{l|ccc|ccc|c}
        \toprule
        \multirow{2}{*}{\textbf{Method}} & \multicolumn{3}{c}{\textbf{Lung Sound Classification}} & \multicolumn{3}{c}{\textbf{Disease Diagnosis}} & \textbf{Metadata} \\
        \cmidrule(lr){2-4} \cmidrule(lr){5-7} \cmidrule(lr){8-8}
        & $S_p$ & $S_e$ & $S_c$ & $S_p$ & $S_e$ & $S_c$ & Accuracy (\%) \\
        \hline
        \midrule
        \multicolumn{8}{c}{\textit{Single Task (Lung or Disease Only)}} \\
        Single Lung Sound & 77.14 $\pm$ 3.35 & 41.97 $\pm$ 2.21 & \underline{59.55} $\pm$ 0.88 & - & - & - & N/A \\
        Single Disease & - & - & - & 74.4 $\pm$ 8.90 & 88.61 $\pm$ 6.77 & 81.51 $\pm$ 2.30 & N/A \\
        \midrule
        \multicolumn{8}{c}{\textit{Two-MTL (Lung + Disease)}} \\
        Hard & 68.93 $\pm$ 6.71 & 45.42 $\pm$ 7.38 & 57.17 $\pm$ 1.40 & 79.32 $\pm$ 5.69 & 89.81 $\pm$ 5.13 & 84.56 $\pm$ 5.01 & N/A \\
        Soft & 70.66 $\pm$ 6.12 & 47.71 $\pm$ 5.07 & 59.19 $\pm$ 0.93 & 82.17 $\pm$ 5.50 & 94.57 $\pm$ 5.38 & 88.37 $\pm$ 2.33 & N/A \\
        \midrule
        \multicolumn{8}{c}{\textit{Tri-MTL (Lung + Disease + Metadata)}} \\
        Hard (Age Group) & 73.89 $\pm$ 1.17 & 40.83 $\pm$ 3.35 & 57.37 $\pm$ 1.91 & 80.23 $\pm$ 4.91 & 93.25 $\pm$ 5.21 & 86.73 $\pm$ 3.68 & 90.29 $\pm$ 3.63 \\
        Hard (Sex) & 73.57 $\pm$ 6.54 & 39.10 $\pm$ 3.12 & 56.33 $\pm$ 2.12 & 76.57 $\pm$ 5.73 & 93.35 $\pm$ 5.78 & 84.96 $\pm$ 7.28 & 51.46 $\pm$ 3.25 \\
        Hard (Location) & 73.27 $\pm$ 6.80 & 42.14 $\pm$ 5.91 & 57.71 $\pm$ 2.19 & 64.23 $\pm$ 4.89 & 96.85 $\pm$ 2.93 & 80.54 $\pm$ 2.11 & 18.04 $\pm$ 2.21 \\
        
        Hard (Stethoscope) & 75.86 $\pm$ 3.32 & 45.28 $\pm$ 3.12 & 58.21 $\pm$ 1.12 & 81.89 $\pm$ 3.42 & 95.82 $\pm$ 4.87 & 88.86 $\pm$ 1.36 & 76.08 $\pm$ 4.25 \\
        \midrule
        Soft (Age Group) & 61.93 $\pm$ 6.33 & 53.67 $\pm$ 7.84 & 57.80 $\pm$ 1.25 & 85.71 $\pm$ 5.35 & 93.32 $\pm$ 2.73 & \underline{89.47} $\pm$ 2.17 & 92.11 $\pm$ 3.12 \\
        Soft (Sex) & 73.07 $\pm$ 6.67 & 39.77 $\pm$ 3.18 & 56.42 $\pm$ 2.37 & 80.91 $\pm$ 5.78 & 94.76 $\pm$ 1.50 & 87.84 $\pm$ 2.72 & 53.06 $\pm$ 3.99 \\
        Soft (Location) & 70.64 $\pm$ 6.37 & 46.15 $\pm$ 6.85 & 58.40 $\pm$ 1.32 & 79.43 $\pm$ 3.32 & 95.52 $\pm$ 1.29 & 87.48 $\pm$ 3.80 & 20.26 $\pm$ 0.47 \\
        Soft (Stethoscope) & 78.86 $\pm$ 7.36 & 41.56 $\pm$ 6.69 & \textbf{60.21} $\pm$ 1.42 & 86.23 $\pm$ 6.50 & 94.09 $\pm$ 0.18 & \textbf{90.16} $\pm$ 3.19 & 81.78 $\pm$ 5.46 \\
        \bottomrule
    \end{tabular}
\end{table*}

\subsubsection{Training Details}
We followed the data pre-processing procedures outlined in previous studies~\cite{gairola2021respirenet, bae23b_interspeech, kim2023adversarial, kim2024stethoscope, 10782363, kim24f_interspeech}, ensuring that all recordings are standardized to 8-second duration with a 16 kHz sampling rate. In our study, all audio feature encoders employed a pretrained AST~\cite{gong2021ast} model and were trained with the Adam optimizer with an initial learning rate of 5e--5 until 50 epochs with a batch size of 8. To mitigate the severe class imbalance in disease diagnosis, we applied a weighted Cross-Entropy loss, where class weights are inversely proportional to the number of samples per class~\cite{gairola2021respirenet, moummad2023pretraining, bae23b_interspeech}. We used 0.1 for $\lambda$ in Equations (\ref{equ6}).

\subsubsection{Metrics}
We evaluated respiratory sound classification and disease diagnosis performance using Specificity ($S_{p}$), Sensitivity ($S_{e}$), and their average Score ($S_c$), adhering to the standardized definitions in~\cite{rocha2018alpha}. To ensure robustness and reliability, every reported value of $S_p$, $S_e$, and $S_c$ reflects the mean and variance across five independent runs where $S_c$ is highest, each initialized with a different random seed (model initialization).

\subsection{Results}
We validate our proposed Tri-MTL framework by comparing its performance against Two-MTL (lung sound classification + disease diagnosis) and single task models. The results presented in Table~\ref{tab:tab2} demonstrates overall experimental results on ICBHI official 60--40\% split dataset.

\subsubsection{Comparison between Single task and Two-MTL}
Single task models serve as a baseline. Lung sound classification as a single task achieves $S_c = 59.55 \pm 0.88$, whereas disease diagnosis achieves $S_c = 81.51 \pm 2.30$. This performance gap indicates that lung sound classification is inherently more challenging. The MTL remarkably improves diagnostic performance compared to single task results. The hard parameter sharing strategy achieves $S_c = 84.56 \pm 6.01$ for disease diagnosis, whereas soft parameter sharing further improves these results, yielding $S_c = 88.37 \pm 2.33$ for disease diagnosis. These findings suggest the substantial benefits of MTL in disease diagnosis, due to the model’s ability to leverage shared representations, which capture complementary features from lung sounds. MTL's lung sound classification task, however, does not similarly benefit from the additional task. This indicates that understanding the diagnosis is not as useful for classifying lung sounds, as understanding the lung sounds is for classifying the diagnosis. However, other additional information such as audio metadata may have similar effect for increasing lung sound classification performance."

\subsubsection{Impact of Metadata Integration in Tri-MTL}
The results of Tri-MTL reveal interesting results in both lung sound classification and disease diagnosis when metadata attributes are incorporated. For the lung sound classification, the Soft (stethoscope) configuration demonstrates the best performance, achieving $S_c = 60.21 \pm 1.42$. This suggests that including recording device-related metadata enables the model to effectively distinguish acoustic variations introduced by different electronic stethoscopes. We confirm that our findings align with the previous research on leveraging stethoscopes can be beneficial for lung sound classification~\cite{kim2024stethoscope, kim24f_interspeech}. For the disease diagnosis performance, similarly, the Soft (stethoscope) configuration achieves the best performance $S_c = 90.77 \pm 1.36$, demonstrating the importance of the stethoscope metadata attribute in improving diagnostic performance.

While including the device information is very useful, metadata classification performance varies across different configurations. The Soft (age group) approach achieves the highest metadata classification accuracy at $92.11 \pm 3.12$, whereas the Hard (location) approach records the lowest accuracy at $18.04 \pm 2.21$. Notably, the Hard (location) configuration also exhibits the worst disease diagnosis performance, falling below that of single task learning. Interestingly, incorporating recording location in the Soft Tri-MTL approach leads to notable improvements in both lung sound classification and disease diagnosis performance compared to Hard Tri-MTL. These findings suggest that certain metadata attributes may contribute positively to model performance.

\begin{figure}[!t]
    \centering
    \begin{subfigure}{.5\linewidth}
      \centering
      \includegraphics[width=1.0\linewidth]{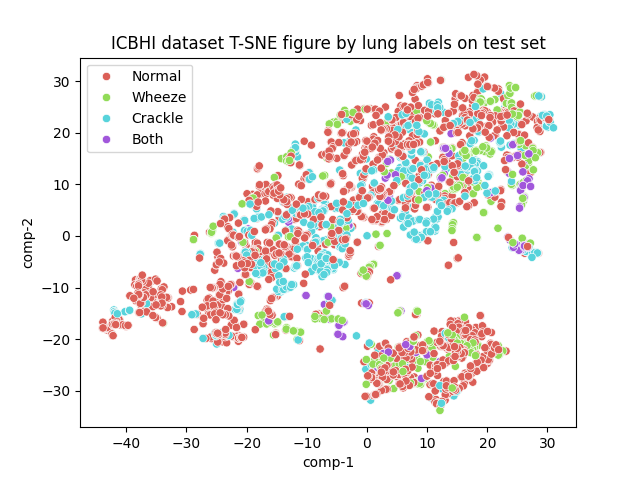}
      \caption{Lung sound classification}
      \label{fig:sfig1}
    \end{subfigure}%
    \begin{subfigure}{.5\linewidth}
      \centering
      \includegraphics[width=1.0\linewidth]{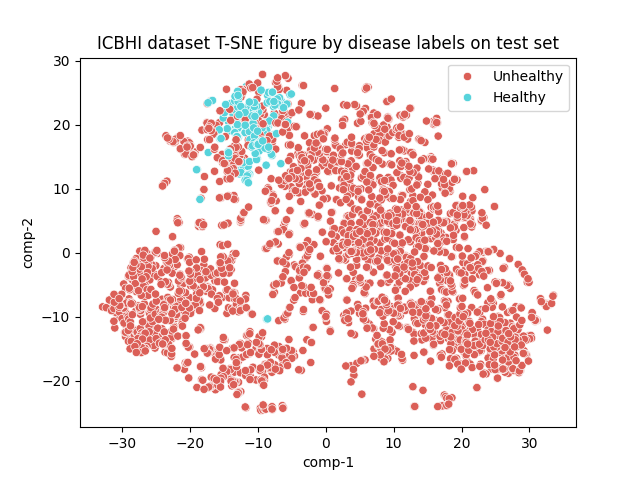}
      \caption{Disease Diagnosis}
      \label{fig:sfig2}
    \end{subfigure}    
    \caption{t-SNE visualizations illustrating how models trained in isolation for a single task organize their feature spaces, highlighting why MTL benefits disease diagnosis more than lung sound classification.}
    \label{fig:fig2}
    \end{figure}

\subsection{Analysis}
To further analyze why MTL improves disease diagnosis but has limited impact on lung sound classification, we utilize t-SNE visualizations of the learned representations. Figure~\ref{fig:sfig1} t-SNE plot illustrates the feature space learned by a model trained exclusively for disease diagnosis, with lung sound labels projected onto this representation. From these visualizations, we notice that the model trained for disease diagnosis struggles to differentiate between various lung sound types (e.g., wheeze, crackle), indicating that while it is optimized for disease classification, it does not inherently learn the fine-grained acoustic characteristics necessary for distinguishing specific lung sound abnormalities.

Conversely, Figure~\ref{fig:sfig2} t-SNE plot visualizes the latent space of a model trained solely for lung sound classification, where disease labels are mapped onto it. Each model operates in isolation, without access to the labels of the other task. In contrast to Figure~\ref{fig:sfig1}, the model trained for lung sound classification effectively captures acoustic features, resulting in a clear distinction between healthy and unhealthy diseases. These findings suggest that MTL can enhance disease diagnosis by leveraging additional information from lung sound labels, thereby improving overall diagnostic performance.

\section{CONCLUSION}
In this study, we introduced Tri-MTL, a multitask learning framework that integrates lung sound classification, disease diagnosis, and metadata attribute classification tasks together to improve individual tasks. We evaluated this approach by exploring both hard and soft parameter sharing approaches for ICHBI respiratory sound dataset. 
Our results demonstrate that incorporating lung sound classification as an auxiliary task enhances disease diagnosis performance, suggesting that shared representations of respiratory sounds contribute to a more accurate diagnostic model. Furthermore, integrating metadata classification, particularly information about the stethoscope type, improves both lung sound classification and disease diagnosis.
These findings underscore the potential clinical impact of multitask learning in respiratory diagnostics.




\bibliography{ref}

\begin{thebibliography}{10}
\providecommand{\url}[1]{#1}
\csname url@samestyle\endcsname
\providecommand{\newblock}{\relax}
\providecommand{\bibinfo}[2]{#2}
\providecommand{\BIBentrySTDinterwordspacing}{\spaceskip=0pt\relax}
\providecommand{\BIBentryALTinterwordstretchfactor}{4}
\providecommand{\BIBentryALTinterwordspacing}{\spaceskip=\fontdimen2\font plus
\BIBentryALTinterwordstretchfactor\fontdimen3\font minus \fontdimen4\font\relax}
\providecommand{\BIBforeignlanguage}[2]{{%
\expandafter\ifx\csname l@#1\endcsname\relax
\typeout{** WARNING: IEEEtran.bst: No hyphenation pattern has been}%
\typeout{** loaded for the language `#1'. Using the pattern for}%
\typeout{** the default language instead.}%
\else
\language=\csname l@#1\endcsname
\fi
#2}}
\providecommand{\BIBdecl}{\relax}
\BIBdecl

\bibitem{coucke2019laennec}
P.~Coucke, ``Laennec versus forbes: tied for the score! how technology helps us interpret auscultation,'' \emph{Revue Medicale de Liege}, vol.~74, no.~10, pp. 543--551, 2019.

\bibitem{sarkar2015auscultation}
M.~Sarkar, I.~Madabhavi, N.~Niranjan, and M.~Dogra, ``Auscultation of the respiratory system,'' \emph{Annals of thoracic medicine}, vol.~10, no.~3, pp. 158--168, 2015.

\bibitem{kim2022coming}
Y.~Kim, Y.~Hyon, S.~Lee, S.-D. Woo, T.~Ha, and C.~Chung, ``The coming era of a new auscultation system for analyzing respiratory sounds,'' \emph{BMC Pulmonary Medicine}, vol.~22, no.~1, p. 119, 2022.

\bibitem{miller2007approach}
C.~J. Miller, ``Approach to the respiratory patient,'' \emph{Veterinary Clinics of North America: Small Animal Practice}, vol.~37, no.~5, pp. 861--878, 2007.

\bibitem{chaichulee2017multi}
S.~Chaichulee, M.~Villarroel, J.~Jorge, C.~Arteta, G.~Green, K.~McCormick, A.~Zisserman, and L.~Tarassenko, ``Multi-task convolutional neural network for patient detection and skin segmentation in continuous non-contact vital sign monitoring,'' in \emph{2017 12th IEEE International Conference on Automatic Face \& Gesture Recognition (FG 2017)}.\hskip 1em plus 0.5em minus 0.4em\relax IEEE, 2017, pp. 266--272.

\bibitem{yang2017novel}
X.~Yang, Z.~Zeng, S.~Y. Yeo, C.~Tan, H.~L. Tey, and Y.~Su, ``A novel multi-task deep learning model for skin lesion segmentation and classification,'' \emph{arXiv preprint arXiv:1703.01025}, 2017.

\bibitem{li2019canet}
X.~Li, X.~Hu, L.~Yu, L.~Zhu, C.-W. Fu, and P.-A. Heng, ``Canet: cross-disease attention network for joint diabetic retinopathy and diabetic macular edema grading,'' \emph{IEEE transactions on medical imaging}, vol.~39, no.~5, pp. 1483--1493, 2019.

\bibitem{liao2019multi}
Q.~Liao, Y.~Ding, Z.~L. Jiang, X.~Wang, C.~Zhang, and Q.~Zhang, ``Multi-task deep convolutional neural network for cancer diagnosis,'' \emph{Neurocomputing}, vol. 348, pp. 66--73, 2019.

\bibitem{amyar2020multi}
A.~Amyar, R.~Modzelewski, H.~Li, and S.~Ruan, ``Multi-task deep learning based ct imaging analysis for covid-19 pneumonia: Classification and segmentation,'' \emph{Computers in biology and medicine}, vol. 126, p. 104037, 2020.

\bibitem{gao2020feature}
F.~Gao, H.~Yoon, T.~Wu, and X.~Chu, ``A feature transfer enabled multi-task deep learning model on medical imaging,'' \emph{Expert Systems with Applications}, vol. 143, p. 112957, 2020.

\bibitem{song2020end}
L.~Song, J.~Lin, Z.~J. Wang, and H.~Wang, ``An end-to-end multi-task deep learning framework for skin lesion analysis,'' \emph{IEEE journal of biomedical and health informatics}, vol.~24, no.~10, pp. 2912--2921, 2020.

\bibitem{suma2024multi}
K.~Suma, D.~Koppad, P.~Kumar, N.~A. Kantikar, and S.~Ramesh, ``Multi-task learning for lung sound and lung disease classification,'' \emph{SN Computer Science}, vol.~6, no.~1, p.~51, 2024.

\bibitem{wang2024multi}
Z.~Wang, H.~Gao, X.~Wang, M.~Grzegorzek, J.~Li, H.~Sun, Y.~Ma, X.~Zhang, Z.~Zhang, A.~Dekker \emph{et~al.}, ``A multi-task learning based applicable ai model simultaneously predicts stage, histology, grade and lnm for cervical cancer before surgery,'' \emph{BMC women's health}, vol.~24, no.~1, p. 425, 2024.

\bibitem{kim2023cross}
S.~Kim, T.~G. Purdie, and C.~McIntosh, ``Cross-task attention network: Improving multi-task learning for medical imaging applications,'' in \emph{International Conference on Medical Image Computing and Computer-Assisted Intervention}.\hskip 1em plus 0.5em minus 0.4em\relax Springer, 2023, pp. 119--128.

\bibitem{kim2023intra}
G.~Kim, H.~Lim, Y.~Kim, O.~Kwon, and J.-H. Choi, ``Intra-person multi-task learning method for chronic-disease prediction,'' \emph{Scientific Reports}, vol.~13, no.~1, p. 1069, 2023.

\bibitem{vandenhende2020revisiting}
S.~Vandenhende, S.~Georgoulis, M.~Proesmans, D.~Dai, and L.~Van~Gool, ``Revisiting multi-task learning in the deep learning era,'' \emph{arXiv preprint arXiv:2004.13379}, vol.~2, no.~3, pp. 5491--5500, 2020.

\bibitem{kv2024multi}
S.~KV, D.~Koppad, P.~Kumar, N.~A. Kantikar, and S.~Ramesh, ``Multi-task learning for lung sound \& lung disease classification,'' \emph{arXiv preprint arXiv:2404.03908}, 2024.

\bibitem{cui2024automated}
S.~Cui and P.~Mitra, ``Automated multi-task learning for joint disease prediction on electronic health records,'' \emph{arXiv preprint arXiv:2403.04086}, 2024.

\bibitem{moummad2023pretraining}
I.~Moummad and N.~Farrugia, ``Pretraining respiratory sound representations using metadata and contrastive learning,'' in \emph{2023 IEEE Workshop on Applications of Signal Processing to Audio and Acoustics (WASPAA)}.\hskip 1em plus 0.5em minus 0.4em\relax IEEE, 2023, pp. 1--5.

\bibitem{kim2024stethoscope}
J.-W. Kim, S.~Bae, W.-Y. Cho, B.~Lee, and H.-Y. Jung, ``Stethoscope-guided supervised contrastive learning for cross-domain adaptation on respiratory sound classification,'' in \emph{ICASSP 2024-2024 IEEE International Conference on Acoustics, Speech and Signal Processing (ICASSP)}.\hskip 1em plus 0.5em minus 0.4em\relax IEEE, 2024, pp. 1431--1435.

\bibitem{kim24f_interspeech}
J.-W. Kim, M.~Toikkanen, Y.~Choi, S.-E. Moon, and H.-Y. Jung, ``Bts: Bridging text and sound modalities for metadata-aided respiratory sound classification,'' in \emph{Interspeech 2024}, 2024, pp. 1690--1694.

\bibitem{kim2025adaptive}
J.-W. Kim, M.~Toikkanen, A.~Jalali, M.~Kim, H.-J. Han, H.~Kim, W.~Shin, H.-Y. Jung, and K.~Kim, ``Adaptive metadata-guided supervised contrastive learning for domain adaptation on respiratory sound classification,'' \emph{IEEE Journal of Biomedical and Health Informatics}, 2025.

\bibitem{gong2021ast}
Y.~Gong, Y.-A. Chung, and J.~Glass, ``{AST: Audio Spectrogram Transformer},'' in \emph{Proc. Interspeech 2021}, 2021, pp. 571--575.

\bibitem{bae23b_interspeech}
S.~Bae, J.-W. Kim, W.-Y. Cho, H.~Baek, S.~Son, B.~Lee, C.~Ha, K.~Tae, S.~Kim, and S.-Y. Yun, ``{Patch-Mix Contrastive Learning with Audio Spectrogram Transformer on Respiratory Sound Classification},'' in \emph{Proc. INTERSPEECH 2023}, 2023, pp. 5436--5440.

\bibitem{rocha2018alpha}
B.~Rocha, D.~Filos, L.~Mendes, I.~Vogiatzis, E.~Perantoni, E.~Kaimakamis, P.~Natsiavas, A.~Oliveira, C.~J{\'a}come, A.~Marques \emph{et~al.}, ``A respiratory sound database for the development of automated classification,'' in \emph{Precision Medicine Powered by pHealth and Connected Health: ICBHI 2017, Thessaloniki, Greece, 18-21 November 2017}.\hskip 1em plus 0.5em minus 0.4em\relax Springer, 2018, pp. 33--37.

\bibitem{ruder2017overview}
S.~Ruder, ``An overview of multi-task learning in deep neural networks,'' \emph{arXiv preprint arXiv:1706.05098}, 2017.

\bibitem{baxter1997bayesian}
J.~Baxter, ``A bayesian/information theoretic model of learning to learn via multiple task sampling,'' \emph{Machine learning}, vol.~28, pp. 7--39, 1997.

\bibitem{caruana1993multitask}
R.~Caruana, ``Multitask learning: A knowledge-based source of inductive bias1,'' in \emph{Proceedings of the Tenth International Conference on Machine Learning}.\hskip 1em plus 0.5em minus 0.4em\relax Citeseer, 1993, pp. 41--48.

\bibitem{duong2015low}
L.~Duong, T.~Cohn, S.~Bird, and P.~Cook, ``Low resource dependency parsing: Cross-lingual parameter sharing in a neural network parser,'' in \emph{Proceedings of the 53rd annual meeting of the Association for Computational Linguistics and the 7th international joint conference on natural language processing (volume 2: short papers)}, 2015, pp. 845--850.

\bibitem{gairola2021respirenet}
S.~Gairola, F.~Tom, N.~Kwatra, and M.~Jain, ``Respirenet: A deep neural network for accurately detecting abnormal lung sounds in limited data setting,'' in \emph{2021 43rd Annual International Conference of the IEEE Engineering in Medicine \& Biology Society (EMBC)}.\hskip 1em plus 0.5em minus 0.4em\relax IEEE, 2021, pp. 527--530.

\bibitem{kim2023adversarial}
\BIBentryALTinterwordspacing
J.-W. Kim, C.~Yoon, M.~Toikkanen, S.~Bae, and H.-Y. Jung, ``Adversarial fine-tuning using generated respiratory sound to address class imbalance,'' in \emph{Deep Generative Models for Health Workshop NeurIPS 2023}, 2023. [Online]. Available: \url{https://openreview.net/forum?id=z1AVG5LDQ7}
\BIBentrySTDinterwordspacing

\bibitem{10782363}
J.-W. Kim, M.~Toikkanen, S.~Bae, M.~Kim, and H.-Y. Jung, ``Repaugment: Input-agnostic representation-level augmentation for respiratory sound classification,'' in \emph{2024 46th Annual International Conference of the IEEE Engineering in Medicine and Biology Society (EMBC)}, 2024, pp. 1--6.

\end{thebibliography}
\bibliographystyle{IEEEtran}

\end{document}